\pdfoutput=1

\documentclass[11pt]{article}

\usepackage[final]{acl}

\newcommand{\aviv}[1]{{\color{RedViolet}[aviv: #1]}}
\newcommand{\hagai}[1]{{\color{blue}[hagai: #1]}}
\newcommand{\yonatan}[1]{{\color{BlueGreen}[yonatan: #1]}}

\newcommand{\dani}[1]{{\color{brown}[Dani: #1]}}

\newcommand{\avivslobodkin}[1]{{\color{red}[avivs: #1]}}

\newcommand{\promptfont}{\fontsize{5.4pt}{6pt}\selectfont}
\renewcommand{\aviv}[1]{}
\renewcommand{\hagai}[1]{}

\renewcommand{\yonatan}[1]{}
\renewcommand{\dani}[1]{}
\renewcommand{\avivslobodkin}[1]{}

\usepackage{graphicx}
\usepackage[normalem]{ulem}
\useunder{\uline}{\ul}{}
\usepackage{makecell}
\usepackage{booktabs, multirow}

\usepackage{arydshln}
\newcommand{\autorater}{\textls[-50]{\textsc{RefVNLI}}}
\newcommand{\BenchmarkCount}{3}
\newcommand{\RefImageName}{\emph{image\textsubscript{ref}}}
\newcommand{\TgtImageName}{\emph{image\textsubscript{tgt}}}
\newcommand{\TextDescName}{\emph{prompt}}

\usepackage{times}
\usepackage{latexsym}

\usepackage[T1]{fontenc}

\usepackage[utf8]{inputenc}

\usepackage{microtype}

\usepackage{inconsolata}

\usepackage{graphicx}

\title{\autorater{}: Towards Scalable Evaluation of \\ Subject-driven Text-to-image Generation}

\author{Aviv Slobodkin$^{1}$\thanks{\hspace{5px}Work done during an internship at Google Research.} \quad Hagai Taitelbaum$^{1}$ \quad Yonatan Bitton$^{1}$ \\ \textbf{Brian Gordon$^{1\ast}$} \quad 
\textbf{Michal Sokolik$^{1}$} \quad \textbf{Nitzan Bitton Guetta$^{2}$} \quad \textbf{Almog Gueta$^{1}$} \\ \textbf{Royi Rassin$^{1\ast}$} \quad \textbf{Dani Lischinski$^{1}$} \quad \textbf{Idan Szpektor$^{1}$} \\  $^{1}$Google Research \qquad $^{2}$Ben Gurion University \\
{\tt\small \{slobodkin,hagait,yonatanbitton\}@google.com}
}

\begin{document}
\maketitle
\begin{abstract}
Subject-driven text-to-image (T2I) generation aims to produce images that align with a given textual description, while preserving the visual identity from a referenced subject image.
Despite its broad downstream applicability---ranging from enhanced personalization in image generation to consistent character representation in video rendering---progress in this field is limited by the lack of reliable automatic evaluation. Existing methods either assess only one aspect of the task (i.e., textual alignment or subject preservation), misalign with human judgments, or rely on costly API-based evaluation.
To address this gap, we introduce \autorater{}, a cost-effective metric that evaluates both textual alignment and subject preservation in a single run.
Trained on a large-scale dataset derived from video-reasoning benchmarks and image perturbations, \autorater{} outperforms or statistically matches existing baselines across multiple benchmarks and subject categories (e.g., \emph{Animal}, \emph{Object}), achieving up to 6.4-point gains in textual alignment and 5.9-point gains in subject preservation.
\end{abstract}

\begin{figure}[t]
  \centering
  \includegraphics[width=0.8\linewidth]{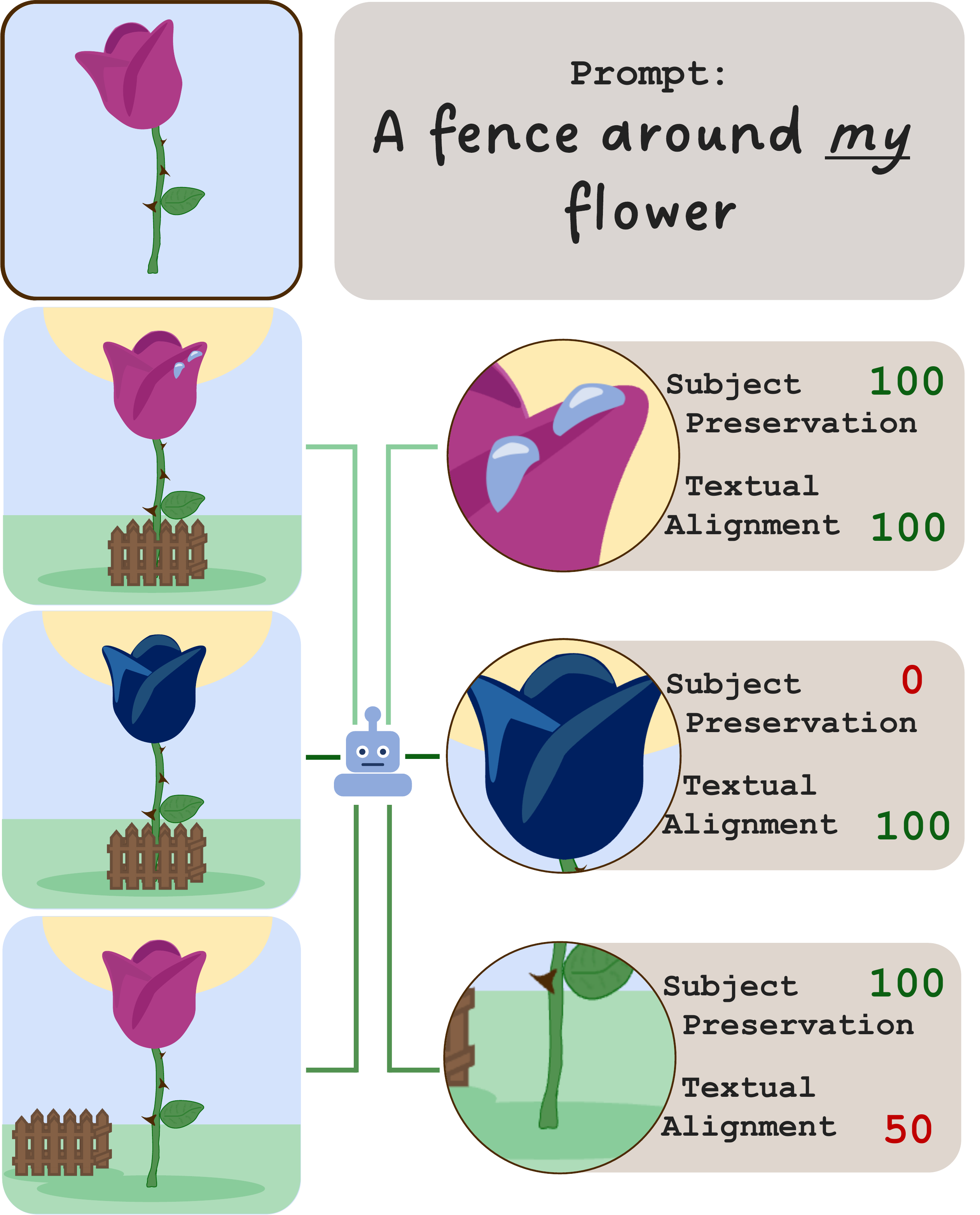}

   \caption{\textbf{Illustration of \autorater{}:} Given a reference image of a subject, a \TextDescName{} referring to the subject, and a target image, \autorater{} assesses both subject preservation and textual alignment. 
   For \textbf{subject preservation}, it distinguishes identity-preserving variations, like dew on a flower (top image), from identity-altering changes, such as color change (middle image).
   For \textbf{textual alignment}, it assesses whether the target image reflects all details from the \TextDescName{}, such as the fence’s position relative to the flower (bottom image).\yonatan{Reminding to explain about small changes that don't change the identity}\aviv{is it more clear?}}
   \label{fig:teaser}
  \vspace{-0.3cm}
\end{figure}

\section{Introduction}
\label{sec:intro}

In a well-known scene from \textit{``The Little Prince''}, the narrator attempts to comfort a grieving prince by saying \textit{``I'll draw you a fence around your flower''}. While fairly simple, this offer raises a deeper question: what makes such a drawing adequate? 
Beyond accurately depicting a fence around a flower, the use of \textit{`your'} implies that it must portray a specific flower---the Prince’s own--- one with which he shares a history. Given the flower’s uniqueness and distinct visual traits, the narrator’s task proves far more complex than it first appears.


\begin{figure*}[t]
  \centering
  \includegraphics[width=0.9\linewidth]{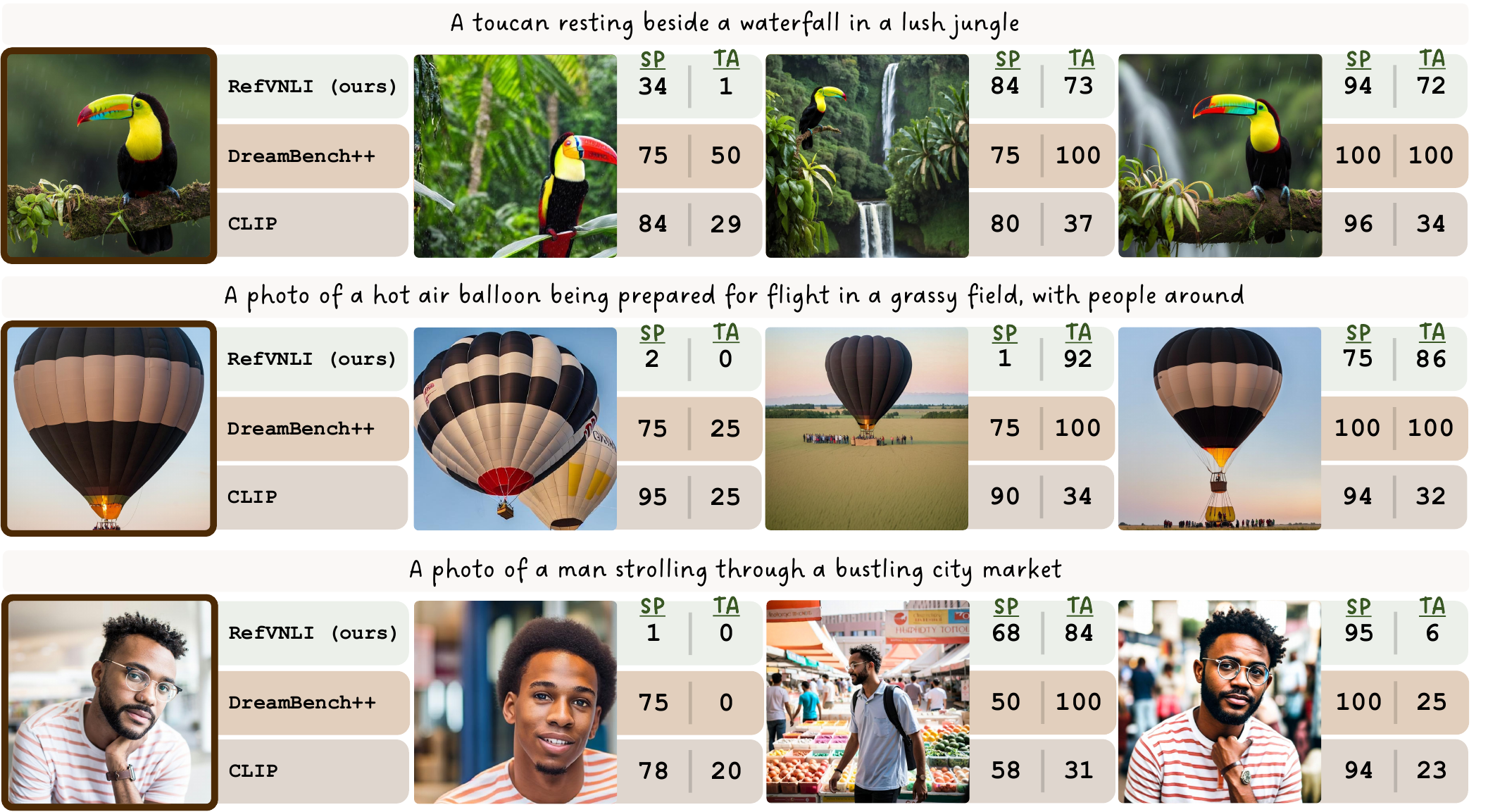}
   \caption{\textbf{Qualitative Comparison}: We compare \autorater{} with DreamBench++ and CLIP, which score both \textbf{Subject Preservation (SP)} and \textbf{Textual Alignment (TA)}, using examples from the \emph{Animal}, \emph{Object}, and \emph{Human} categories.  DreamBench++ scores (0-4) are scaled to 0-100 for better readability. 
   \autorater{} exhibits better robustness to identity-agnostic changes (SP), such as the zoomed-out parrot (top-middle) and the zoomed-out person with different attire (bottom-middle).  It is also more sensitive to identity-defining traits, penalizing changed facial features (left-most person) and mismatched object patterns (left and middle balloons).    
   Additionally, \autorater{} excels at detecting text-image mismatches (TA), as seen in its penalization of the top-left image for lacking a waterfall.}
   \label{fig:qualitative_examples_text_and_image}
\end{figure*}

Subject-driven text-to-image (T2I) generation \cite{chen2024subject, li2024blip, ruiz2023dreambooth} enables a variety of downstream applications, such as personalized image generation \cite{ruiz2023dreambooth}, character consistency in video generation \cite{liu2024evalcrafter}, and enhancing vanilla T2I evaluation frameworks for less-known entities via image retrieval \citep{tahmasebi2025verifying}.
Unlike standard T2I models that are only conditioned on text inputs, this setup takes both a textual prompt and a reference image, enabling more precise subject representation.
For example, when creating an image of a fenced-in flower for the Little Prince, subject-driven models should use a reference image of the Prince's flower to ensure the output preserves its unique features (see Fig.~\ref{fig:teaser}).

Despite its wide applicability, research in this area has been hindered by a lack of reliable and scalable auto-raters.
Existing metrics typically correlate poorly with humans, and often focus only on \textit{textual alignment} between the input prompt and the target image, as in CLIP-T \cite{radford2021learning} and SigLIP \cite{zhai2023sigmoid}, or on \textit{subject preservation} between the input and target images, as in CLIP-I \cite{radford2021learning} and DINO-I \cite{caron2021emerging}, while both aspects are needed for successful subject-driven generation. 
More correlative metrics, like DreamBench++ \cite{peng2024dreambenchhumanalignedbenchmarkpersonalized} and VIEScore \cite{ku-etal-2024-viescore}, depend on costly API calls to models like GPT-4 \cite{openai2024gpt4technicalreport}, making them less scalable and reproducible. \avivslobodkin{added the "making them less scalable and reproducible."}

\hagai{do we want to add about them attending only over one of the inputs and not over both of them together?}
To bridge this gap, we present \autorater{}, a cost-effective fine-tuned auto-rater for subject-driven T2I generation. Given a triplet $<$\RefImageName{}, \TextDescName{}, \TgtImageName{}$>$, \autorater{} predicts two scores—\emph{textual alignment} and \emph{subject preservation}—in a single run, as shown in Fig.~\ref{fig:teaser}.
To train \autorater{}, we automatically curate a large-scale dataset of $<$\RefImageName{}, \TextDescName{}, \TgtImageName{}$>$ triplets, labeled with \textit{$<$textual alignment, subject preservation$>$ $\in\{0,1\}^2$}. 
For subject preservation, we 
identify subjects across video frames, creating positive examples using pairs of frames depicting the same subject, and negative ones by pairing frames of different subjects (yet of the same entity).
This approach enables robustness to variations in subject appearance (e.g., rotation, setting, clothing), as well as to the presence of extraneous elements (e.g., dew on the Little Prince’s flower in Fig.~\ref{fig:teaser}, top).
At the same time, \autorater{} must also be sensitive to identity-defining traits, such as human facial features or object shapes and colors (e.g., middle image in Fig.~\ref{fig:teaser}).
To this end, 
we modify images by masking and inpainting identity-critical regions, while keeping everything else unchanged.
The original subject crops are then paired with the unaltered images as positive pairs and with the modified images as negative pairs, thereby teaching the model to focus on key identity attributes.

\begin{figure*}[t]
  \centering
  \includegraphics[width=0.8\textwidth]{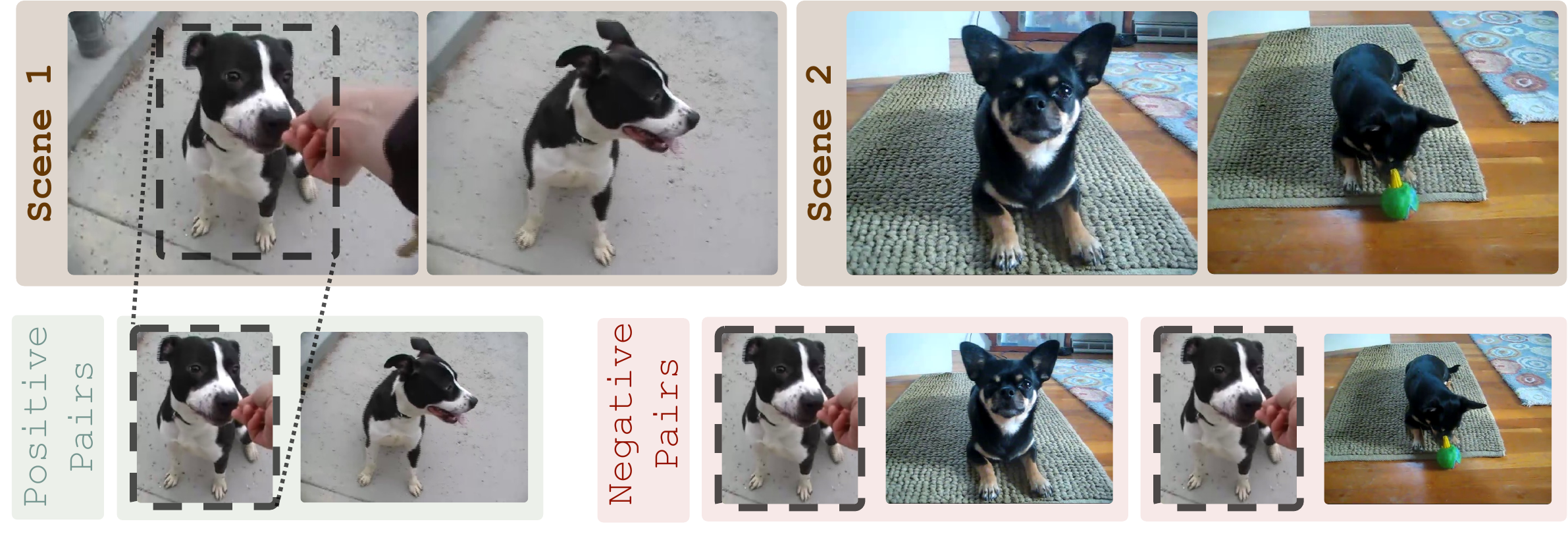}

   \caption{Generating subject preservation classification training instances from video frames. Given two pairs of frames, each extracted from distinct video scenes featuring the same entity (e.g., \textit{a} dog), where both frames within each pair depict the same subject (e.g., \textit{the same} dog), we construct training \{\RefImageName{}, \TgtImageName{}\} pairs for subject preservation classification.
   \textbf{Positive pairs} are formed by pairing a cropped subject from one frame (e.g., dog from left frame in Scene 1) with the full frame from the same scene (right frame in Scene 1). In contrast, \textbf{negative pairs} are created by pairing the cropped subject with the other scene's full frames (e.g., Scene 2). This process is applied to all four frames, with each taking turns as the cropped reference image (\RefImageName{}), while the corresponding full-frame counterparts serve as \TgtImageName{}, yielding a total of 4 positive and 8 negative training pairs.}
   \label{fig:scarping_videos}
\end{figure*}

For textual alignment, we first create positive \emph{image}-\TextDescName{} pairs.
For that, we use an LLM
to caption each image in the aforementioned pairs, ensuring focus on the subject by enclosing it within a bounding box. Negative pairs are then formed by replacing these captions with those of different scenes. For extra sensitivity to minor mismatches, like a fence drawn \textit{next to} rather than \textit{around} the Prince’s flower (Fig.~\ref{fig:teaser}, bottom), we also create negative pairs by altering a single fact in each original (positive) caption.
Finally, to derive the $<$\RefImageName{}, \TextDescName{}, \TgtImageName{}$>$ triplets from each pair of frames and associated captions, we use the cropped subject from one frame as \RefImageName{} and the entire second frame, alongside its caption, as \{\TextDescName{}, \TgtImageName{}\}, resulting in a total of 1.2 million instances.


We evaluate \autorater{} on multiple human-labeled test sets for subject-driven generation, including DreamBench++ \cite{peng2024dreambenchhumanalignedbenchmarkpersonalized}, ImagenHub \cite{ku2310imagenhub}, and KITTEN \cite{huang2024kittenknowledgeintensiveevaluationimage}, across categories such as \emph{Humans}, \emph{Animals}, \emph{Objects}, \emph{Landmarks}, and a multi-subject setting.
For \textbf{textual alignment}, \autorater{} consistently matches or outperforms all baselines, with up to 6.4-point gains in \emph{Landmarks} and proficiency at detecting subtle text-image misalignments (e.g., missing waterfall in Fig.~\ref{fig:qualitative_examples_text_and_image}, top-left).
It also leads in \textbf{subject preservation}, with gains of up to 6.3 points on \emph{Objects} (Table~\ref{tab:dreambench_results}) and 5.9 points in the multi-subject setting, surpassing the larger GPT-4o-based DreamBench++ baseline.
As seen in Fig.~\ref{fig:qualitative_examples_text_and_image}, it balances robustness to non-critical changes (e.g., zoomed-out toucan, top-middle) with sensitivity to identity shifts (e.g., altered facial features, bottom-left).
Further, \autorater{} effectively handles rare subjects (\S\ref{sec:unknown_entities}), 
outperforming all baselines and highlighting its value as a reliable alternative to standard T2I metrics for uncommon entities. \avivslobodkin{To Idan: how is this alternative ending?}

\section{\autorater{}: Automatic Metric for Subject-driven T2I Generation}
\label{sec:refvnli}

We introduce \autorater{}, a cost-effective auto-rater specifically tailored for subject-driven T2I generation.
This section details the automated pipeline used to construct its training dataset (\S\ref{subsec:trainset_construction}) and the subsequent training process of \autorater{} (\S\ref{subsec:refvnli_training}).


\subsection{Training Dataset Construction}
\label{subsec:trainset_construction}

To train \autorater{}, we collect a large scale dataset of $<$\RefImageName{}, \TextDescName{}, \TgtImageName{}$>$ triplets, each with two binary labels: one for \textbf{subject preservation} of \RefImageName{} in \TgtImageName{}, and one for \textbf{textual alignment} between the \TextDescName{} and \TgtImageName{}.
This involves first creating subject-driven \{\RefImageName{}, \TgtImageName{}\} pairs, followed by automatic generation of subject-focused \TextDescName{}\emph{s} for each \TgtImageName{}.

\begin{figure*}[t!]
  \centering
  \includegraphics[width=0.85\textwidth]{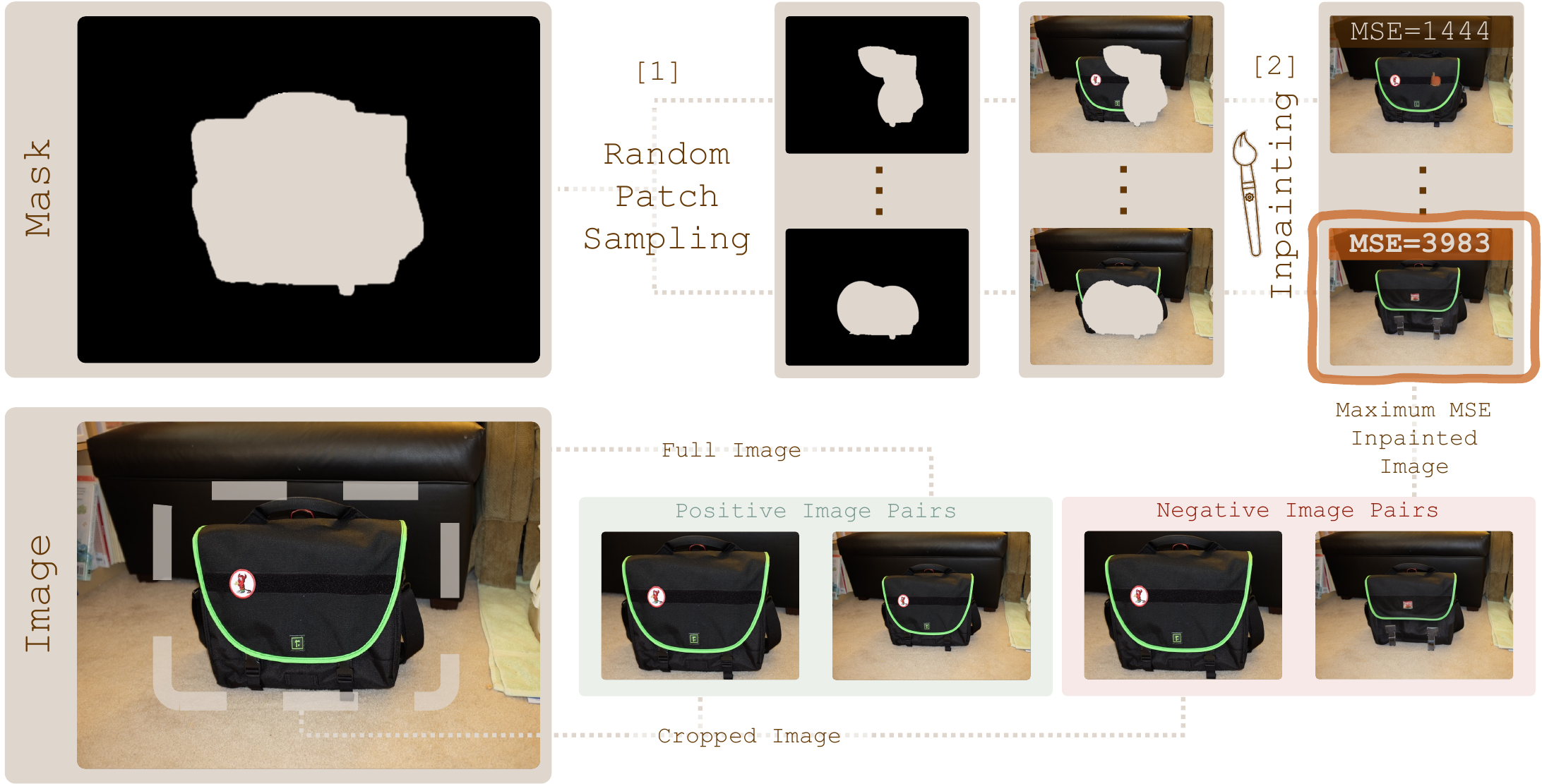}

   \caption{Creating identity-sensitive \{\RefImageName{}, \TgtImageName{}\} pairs.
   Starting with an image and a mask of a subject (e.g., a bag), we randomly keep 5 patches within the masked area ([1]) and use them to create 5 inpainted versions ([2]). The version with the highest MSE between the altered and original areas (e.g., bottom image, MSE = 3983) is paired with the \textit{unmodified} crop to form a \textbf{negative pair}, while the original image and the same crop create a \textbf{positive pair}, with the crop acting as \RefImageName{} in both cases.}
   \label{fig:corrupting_images}
\end{figure*}

\paragraph{Subject-driven image pairs.}
\label{subsec:trainset_construction_subject_consistency}
To ensure our \{\RefImageName{}, \TgtImageName{}\} dataset is robust to identity-agnostic changes (e.g., pose, clothing, or lighting changes), we use video-based datasets that inherently capture these differences. Specifically, we use Mementos \cite{wang-etal-2024-mementos}, comprising scene-specific video frames with human-written textual descriptions, and TVQA+ \cite{lei-etal-2020-tvqa}, containing human-annotated bounding boxes for characters and objects in TV episodes.
We first locate subjects within frames: for Mementos, we extract entities from the provided textual descriptions using Gemini \cite{geminiteam2024geminifamilyhighlycapable} and localize them in the associated frames with an object detection model \cite{minderer2022simple}, while for TVQA+, we directly use the provided bounding boxes. 
\textbf{Positive pairs} are formed from frames featuring the same subject, usually within the same scene,\footnote{For TVQA+, we also include cross-scene positive pairs for named entities, such as TV characters.} while \textbf{negative pairs} consist of frames with distinct subjects (of the same type of entity), often across scenes (see Fig.~\ref{fig:scarping_videos}). 
These \textit{frame}-pairs are then converted into \{\RefImageName{}, \TgtImageName{}\}-pairs by cropping the subject from one frame as \RefImageName{} (e.g., left frame in Fig.~\ref{fig:scarping_videos}, Scene 1) and using the full second frame as \TgtImageName{} (each of the other frames in Fig.~\ref{fig:scarping_videos}). This is then repeated with reversed roles for an extra \{\RefImageName{}, \TgtImageName{}\} pair.
In total, we collected 338,551 image pairs (228,661 from Mementos and 109,890 from TVQA+) from 44,418 unique frames.

To further enhance sensitivity to identity-specific attributes, such as facial features in humans or shapes and patterns in objects, we leverage the Open Images dataset \cite{kuznetsova2020open} to create additional training instances, as shown in Fig.~\ref{fig:corrupting_images}. Using its gold segmentation masks, we selectively mask and inpaint identity-critical regions while preserving other details. Specifically, we randomly sample 5 sub-masks covering 30\%-50\% of the subject mask ([1] in Fig.~\ref{fig:corrupting_images}), which we use to create 5 inpainted variants ([2]). 
The version with the highest Mean Squared Error (MSE) between the modified and original regions (e.g., Fig.~\ref{fig:corrupting_images}, bottom image, MSE=3983) is then paired with the \textit{unmodified} cropped subject to form a \textbf{negative pair} of \{\RefImageName{}, \TgtImageName{}\}, while the original image and the same crop form a \textbf{positive pair}, with the crop serving as \RefImageName{} in both cases.
This process yields extra 16,572 pairs, helping the model focus on fine-grained identity details.
To further improve data quality, we also apply multiple filtering steps, including removing blurry images and those with unclear subjects (see Appendix~\ref{subappendix:collection_of_subject_drive_image_pairs} for details).

\begin{figure}[t!]
  \centering
  \includegraphics[width=\linewidth]{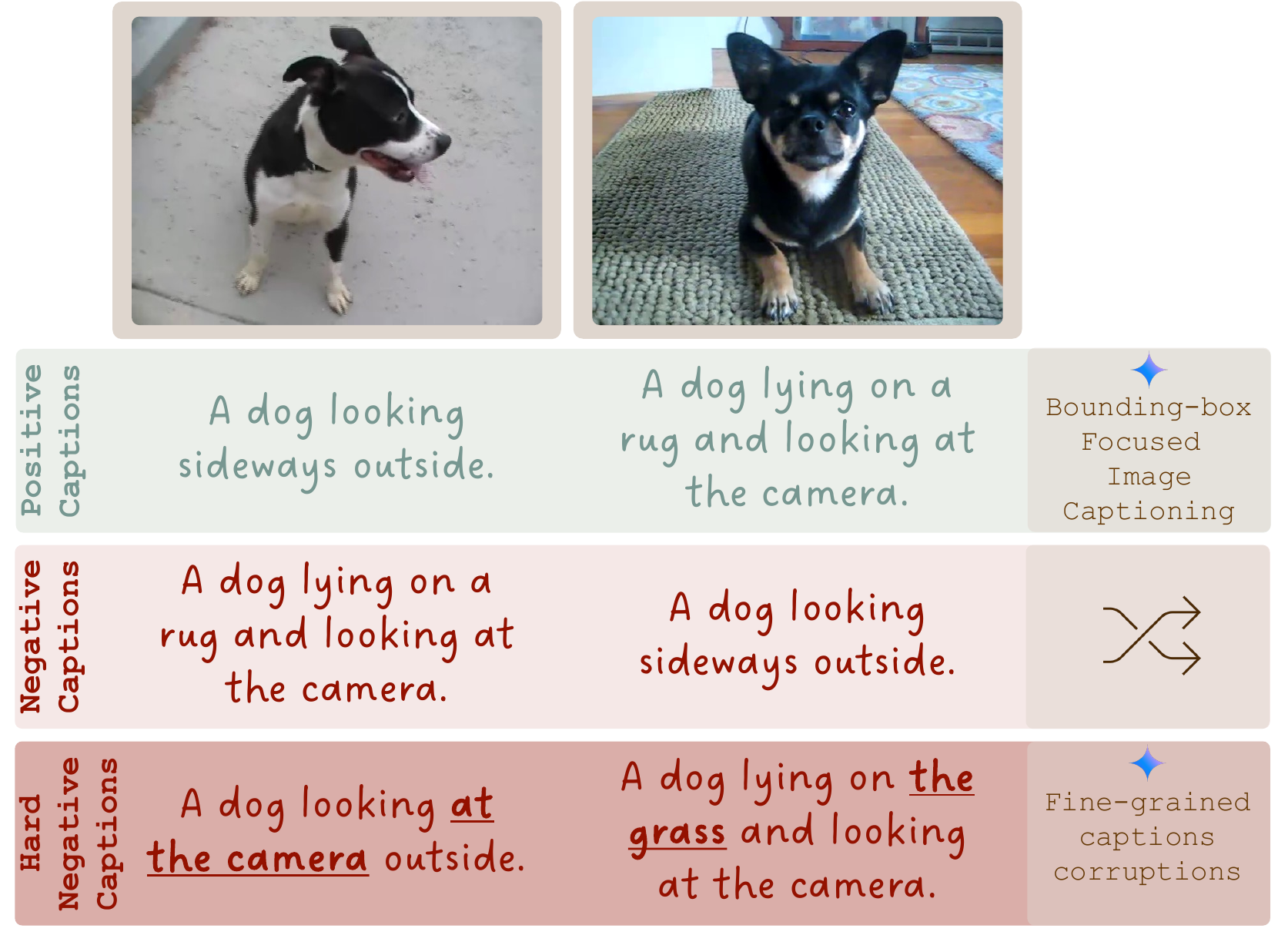}

   \caption{Example of \TextDescName{}-\TgtImageName{} pairs. Given an image with some subject (e.g., a dog), we create a \textbf{positive \TextDescName{}} by adding a bounding box around the subject and directing Gemini to describe it (top \TextDescName{}s). \textbf{Negative \TextDescName{}s} are created by swapping \TextDescName{}s between images of the same entity (middle \TextDescName{}s). For additional \textbf{hard negatives}, we guide Gemini to modify a single non-subject detail in the positive \TextDescName{} while keeping the rest unchanged (bottom \TextDescName{}s).}
   \label{fig:caption_generation}
  \vspace{-0.25cm}
\end{figure}

\paragraph{Image-\TextDescName{} pairs.}
\label{subsec:trainset_construction_textual_alignment}
For each \{\RefImageName{}, \TgtImageName{}\} pair, we generate positive and negative \TextDescName{}\emph{s} for \TgtImageName{} (Fig.~\ref{fig:caption_generation}). 
\textbf{Positive \TextDescName{}\emph{s}} (Fig.~\ref{fig:caption_generation}, top) are created by instructing Gemini \cite{geminiteam2024geminifamilyhighlycapable} to describe \TgtImageName{}, ensuring the subject is explicitly mentioned by enclosing it in a bounding box and guiding the model to focus on it, as well as filtering out \TextDescName{}\emph{s} lacking it.
For \textbf{negative \TextDescName{}\emph{s}} (Fig.~\ref{fig:caption_generation}, middle), we swap \TextDescName{}\emph{s} between frames containing the same entity type (e.g., a dog).
To further enhance sensitivity to subtle mismatches, we also create \textbf{hard-negative \TextDescName{}\emph{s}} (Fig.~\ref{fig:caption_generation}, bottom) by using Gemini to modify a single non-subject detail in the positive \TextDescName{}\emph{s}, following \citet{gordon2024mismatch}. 
In total, this and the image-pairing steps yield 1.2 million $<$\RefImageName{}, \TextDescName{}, \TgtImageName{}$>$ triplets labeled for \emph{textual alignment} and \emph{subject preservation}.


\subsection{\autorater{} Training}
\label{subsec:refvnli_training}
We fine-tune PaliGemma \cite{beyer2024paligemma}, a 3B Vision-Language Model (VLM) known for effective transfer learning, focusing on a variant adapted for multi-image inputs.\footnote{\url{https://huggingface.co/google/paligemma-3b-ft-nlvr2-448}} 
The model takes as input two images (\RefImageName{} and \TgtImageName{}), and a \TextDescName{} that includes \textit{$<$u$>$} and \textit{$<$\textbackslash{}u$>$} markups around the referenced subject.
During training, the model performs two sequential binary classifications---first assessing textual alignment, then subject preservation---outputting \textit{`1'} (positive) or \textit{`0'} (negative) for each task. 
At inference, we compute the probabilities of predicting \textit{`1'} and \textit{`0'} for the first and second generated tokens, and use their ratio to calculate the textual alignment and subject preservation scores, respectively.
\footnote{See Appendices~\ref{appendix:Reproducibility} and \ref{appendix:ablations} for more details and ablations.}
\aviv{time-permitting, add an illustration of this model and all ablations in the appendix.}

\section{Experimental Settings}
\label{sec:experimental_setting}
This section outlines our meta-evaluation protocol and benchmarks (\S\ref{subsec:meta_eval_and-benchmarks}), followed by an overview of the baseline models used for comparison (\S\ref{subsec:experiment_setting_baselines}).

\subsection{Meta-evaluation and Benchmarks}
\label{subsec:meta_eval_and-benchmarks}

We include \BenchmarkCount{} subject-driven generation benchmarks with human annotations for textual alignment and subject preservation across categories such as \emph{Human}, \emph{Animal}, \emph{Object}, and \emph{Landmark}. 
To enable a unified evaluation framework, given differing scoring methods (5-scale and binary), we convert all annotations into binary labels: one for whether \TgtImageName{} fully captures the \TextDescName{} (textual alignment) and another for whether it correctly depicts the referenced subject (subject preservation). 

For meta-evaluation, we report ROC AUC for each criterion, following standard practice \cite{honovich2022true, yarom2023you, zha2023alignscore}, and also compute a unified score as the harmonic mean of the two scores. Following \citet{honovich2022true}, significance testing is assessed via bootstrap resampling \citep{efron1987better}, comparing each baseline to \autorater{}. We report mean scores and highlight models with statistically significant under- or outperformance relative to \autorater{}.

We next present the \BenchmarkCount{} analyzed benchmarks.

\paragraph{Dreambench++} \cite{peng2024dreambenchhumanalignedbenchmarkpersonalized} is a subject-driven generation benchmark with human annotations for 8,190 images generated by 7 models. Annotators rated textual alignment and subject preservation on a 0-4 scale, with each image evaluated by 2 raters. 
To convert these ratings into binary labels, we classify a criterion as positive if both scores are at least 3, and at least one is a 4.
We report performance separately for the benchmark’s three subject categories: \textbf{Human}, \textbf{Animal}, and \textbf{Object}.\footnote{A fourth `style' category is excluded as it is beyond our work's scope.}

\begin{table*}[!htp]
\centering
\resizebox{0.8\textwidth}{!}{%
\begin{tabular}{lccc@{\hskip 20pt}ccc@{\hskip 20pt}cccc}\toprule
&\multicolumn{3}{c}{\textbf{Textual Alignment}}
{\hskip 5pt}
&\multicolumn{3}{c}{\textbf{Subject Preservation}}
{\hskip 16pt}
&\multicolumn{3}{c}{\textbf{Unified Evaluation}}
{\hskip 1pt} \\
&Animal &Human &Object &Animal &Human &Object &Animal &Human &Object \\\midrule
CLIP &\hspace{1mm} 72.8$^{\downarrow}$ &\hspace{1mm} 77.4$^{\downarrow}$ &\hspace{1mm} 74.6$^{\downarrow}$ &\hspace{1mm} 72.4$^{\downarrow}$ &\textbf{87.7} &\hspace{1mm} 76.4$^{\downarrow}$ &\hspace{1mm} 72.6$^{\downarrow}$ &82.2 &\hspace{1mm} 75.5$^{\downarrow}$ \\
DINO &- &- &- &\textbf{80.1} &\hspace{1mm} 78.0$^{\downarrow}$ &\hspace{1mm} 77.3$^{\downarrow}$ &- &- &- \\
Crop-IR &- &- &- &\hspace{1mm} 76.9$^{\downarrow}$ &85.6 &83.4 &- &- &- \\
ArcFace &- &- &- &- &\hspace{1mm} 61.0$^{\downarrow}$ &- &- &- &- \\
CLIPScore &\hspace{1mm} 71.5$^{\downarrow}$ &\hspace{1mm} 76.1$^{\downarrow}$ &\hspace{1mm} 72.9$^{\downarrow}$ &- &- &- &- &- &- \\
BLIPScore &\hspace{1mm} 75.4$^{\downarrow}$ &\hspace{1mm} 79.5$^{\downarrow}$ &\hspace{1mm} 78.9$^{\downarrow}$ &- &- &- &- &- &- \\
SigLIP &\hspace{1mm} 72.5$^{\downarrow}$ &80.2 &\hspace{1mm} 77.1$^{\downarrow}$ &- &- &- &- &- &- \\
TIFA &\hspace{1mm} 70.6$^{\downarrow}$ &\hspace{1mm} 75.7$^{\downarrow}$ &\hspace{1mm} 69.5$^{\downarrow}$ &- &- &- &- &- &- \\
VQAScore &79.4 &\hspace{1mm} 78.0$^{\downarrow}$ &\textbf{82.6} &- &- &- &- &- &- \\
VIEScore &77.9 &\hspace{1mm} 75.2$^{\downarrow}$ &\hspace{1mm} 73.3$^{\downarrow}$ &\hspace{1mm} 63.4$^{\downarrow}$ &81.1 &\hspace{1mm} 76.4$^{\downarrow}$ &\hspace{1mm} 69.8$^{\downarrow}$ &77.7 &\hspace{1mm} 74.8$^{\downarrow}$ \\
DreamBench++ &79.5 &\textbf{82.7} &82.5 &\hspace{1mm} 74.5$^{\downarrow}$ &84.1 &\hspace{1mm} 79.4$^{\downarrow}$ &\hspace{1mm} 76.9$^{\downarrow}$ &83.4 &\hspace{1mm} 80.9$^{\downarrow}$ \\
\hdashline
PaliGemma\textsubscript{text/ref} &\hspace{1mm} 77.9$^{\downarrow}$ &\hspace{1mm} 79.2$^{\downarrow}$ &81.2 &\hspace{1mm} 70.1$^{\downarrow}$ &\hspace{1mm} 71.2$^{\downarrow}$ &\hspace{1mm} 77.6$^{\downarrow}$ &\hspace{1mm} 73.8$^{\downarrow}$ &\hspace{1mm} 74.9$^{\downarrow}$ &\hspace{1mm} 79.4$^{\downarrow}$ \\
\autorater{} &\textbf{80.2} &82.5 &82.0 &79.4 &86.0 &\textbf{85.7} &\textbf{79.8} &\textbf{84.2} &\textbf{83.8} \\
\bottomrule
\end{tabular}%
}
\caption{\textbf{ROC AUC scores on DreamBench++} for textual alignment, subject preservation, and their harmonic mean (as a unified evaluation) across \emph{Animal}, \emph{Human}, and \emph{Object} categories. 
The last two rows feature models finetuned on our dataset, with PaliGemma\textsubscript{text/ref} comprising two separate models (PaliGemma\textsubscript{text} and PaliGemma\textsubscript{ref}) trained exclusively for each criterion. Bold indicates the highest score per column. $^{\downarrow}$ and $^{\uparrow}$ indicate statistically significant underperformance and outperformance relative to \autorater{}, respectively.}
\label{tab:dreambench_results}
\end{table*}

\paragraph{ImagenHub} 
\citep{ku2310imagenhub} is a human-annotated benchmark for conditional image generation, covering a subject-driven task (150 instances) and a multi-concept task (102 instances), which involves 2 referenced subjects per instance. 
Each image was rated by 3 annotators.
Instead of separate ratings for textual alignment and subject preservation, annotators provided a single adherence score (0, 0.5, or 1) per image. To align with our binary labeling framework, images rated 1 by all 3 annotators were assigned positive labels for both criteria, while the rest were re-annotated by this paper's authors. In the Multi-subject setting, a positive subject preservation label was assigned only when both subjects were accurately depicted.
For evaluation, we report separate scores for \textbf{Animals} and \textbf{Objects} in the single-subject task, while \textbf{Multi-subject} instances are split into two single-subject evaluations, with the final score being the lower rating (per criterion), to ensure a stricter assessment.


\paragraph{KITTEN} \cite{huang2024kittenknowledgeintensiveevaluationimage} evaluates subject-driven T2I models on generating diverse real-world entities (e.g., plants, vehicles, landmarks), using 5 reference images and a \TextDescName{}.
Annotators rated entity depiction on a 1–5 scale and provided binary textual alignment scores, with each image assessed by 5 annotators.
Unlike our focus on \textit{specific subjects}, KITTEN evaluates \textit{general entity alignment} (e.g., a generic rose rather than a specific one). Hence, we only use the 256 \textbf{Landmark} images, as landmarks are unique entities where \textit{entity} adherence coincides with \textit{subject} adherence. 
To convert ratings into binary labels, we apply majority voting for textual alignment and consider subject preservation positive only if most annotators rated it at least 4 and the average score is 4 or higher.


\subsection{Baselines}
\label{subsec:experiment_setting_baselines}
We evaluate \autorater{} against both standard and state-of-the-art methods for measuring textual alignment, subject preservation, or both.

\paragraph{Baselines for textual alignment.}
We compare \autorater{} with two groups of automatic metrics for textual alignment. The first group leverages large vision-language models (VLMs), computing cosine similarity between text and image encodings. This includes BLIPScore \cite{li2022blip}, CLIPScore \cite{hessel2021clipscore}, and SigLIP \cite{zhai2023sigmoid}. The second group, which includes TIFA \cite{hu2023tifa} and VQAScore \cite{lin2024evaluating}, evaluates textual alignment via visual question answering (VQA). 
We also include a baseline where PaliGemma is finetuned on our dataset exclusively for textual alignment, given only the prompt and target image, referred to as PaliGemma\textsubscript{text}.

\paragraph{Baselines for subject preservation.}
For subject preservation, we compare \autorater{} to baselines that use large VLMs by computing cosine similarity between reference and target image embeddings. These include DINO \cite{caron2021emerging}, Crop-IR \cite{winter2024objectmate},\footnote{See Appendix~\ref{appendix:crop_ir_object_detection_ablation} for more details.} and for the \emph{Human} category, also ArcFace \cite{Deng_2019_CVPR}, a face-recognition model.
We also assess a PaliGemma model finetuned on our dataset solely for subject preservation, using only reference and target images (formatted as in \S\ref{subsec:refvnli_training}), denoted as PaliGemma\textsubscript{ref}.

\paragraph{Baselines for both criteria.}
We also include 3 metrics that assess both criteria. 
CLIP \cite{radford2021learning} computes scores separately for each criterion by calculating cosine similarity between the encodings of \TgtImageName{} and those of \TextDescName{} and \RefImageName{}.
VIEScore \cite{ku-etal-2024-viescore} uses an elaborate GPT-4o \cite{openai2024gpt4technicalreport} few-shot strategy, simultaneously generating two 0–10 ratings, one for each criterion. 
Lastly, DreamBench++ \cite{peng2024dreambenchhumanalignedbenchmarkpersonalized} evaluates each criterion separately using distinct GPT-4o prompts with hand-crafted instructions and examples. This method follows a two-step prompting process, where GPT-4o first summarizes the evaluation task to increase task comprehension before assigning a 0–4 score.

\begin{table*}[!t]
\centering
\resizebox{0.85\textwidth}{!}{%
\begin{tabular}{lccc@{\hskip 20pt}ccc@{\hskip 20pt}cccc}\toprule
&\multicolumn{3}{c}{\textbf{Textual Alignment}} {\hskip 15pt} &\multicolumn{3}{c}{\textbf{Subject Preservation}} {\hskip 20pt} &\multicolumn{3}{c}{\textbf{Unified Evaluation}} {\hskip 20pt} \\
\addlinespace[2pt]
&Animal &Object &Multi-subj. &Animal &Object &Multi-subj. &Animal &Object &Multi-subj. \\\midrule
CLIP &81.8 &\hspace{1mm} 74.7$^{\downarrow}$ &\hspace{1mm} 81.1$^{\downarrow}$ &\hspace{1mm} 63.8$^{\downarrow}$ &\hspace{1mm} 73.3$^{\downarrow}$ &\hspace{1mm} 52.6$^{\downarrow}$ &\hspace{1mm} 71.6$^{\downarrow}$ &\hspace{1mm} 74.0$^{\downarrow}$ &\hspace{1mm} 63.8$^{\downarrow}$ \\
DINO &- &- &- &81.7 &\hspace{1mm} 77.3$^{\downarrow}$ &\hspace{1mm} 50.0$^{\downarrow}$ &- &- &- \\
Crop-IR &- &- &- &77.6 &\textbf{84.1} &56.8 &- &- &- \\
CLIPScore &81.5 &\hspace{1mm} 75.0$^{\downarrow}$ &\hspace{1mm} 79.1$^{\downarrow}$ &- &- &- &- &- &- \\
BLIPScore &82.9 &\hspace{1mm} 79.7$^{\downarrow}$ &\hspace{1mm} 84.2$^{\downarrow}$ &- &- &- &- &- &- \\
SigLIP &80.7 &\hspace{1mm} 80.6$^{\downarrow}$ &\hspace{1mm} 82.3$^{\downarrow}$ &- &- &- &- &- &- \\
TIFA &79.9 &\hspace{1mm} 76.1$^{\downarrow}$ &\hspace{1mm} 79.2$^{\downarrow}$ &- &- &- &- &- &- \\
VQAScore &\hspace{1mm} 77.3$^{\downarrow}$ &\hspace{1mm} 83.8$^{\downarrow}$ &87.8 &- &- &- &- &- &- \\
VIEScore &\hspace{1mm} 62.1$^{\downarrow}$ &\hspace{1mm} 54.1$^{\downarrow}$ &\hspace{1mm} 71.6$^{\downarrow}$ &\hspace{1mm} 56.4$^{\downarrow}$ &\hspace{1mm} 49.4$^{\downarrow}$ &\hspace{1mm} 50.2$^{\downarrow}$ &\hspace{1mm} 59.0$^{\downarrow}$ &\hspace{1mm} 51.5$^{\downarrow}$ &\hspace{1mm} 58.9$^{\downarrow}$ \\
DreamBench++ &\textbf{86.4} &\hspace{1mm} 85.5$^{\downarrow}$ &\textbf{88.2} &\hspace{1mm} 71.1$^{\downarrow}$ &84.0 &\hspace{1mm} 54.3$^{\downarrow}$ &\hspace{1mm} 78.0$^{\downarrow}$ &84.8 &\hspace{1mm} 67.2$^{\downarrow}$ \\
\hdashline
PaliGemma\textsubscript{text/ref} &81.1 &88.1 &85.3 &\textbf{82.0} &\hspace{1mm} 74.2$^{\downarrow}$ &62.1 &81.5 &\hspace{1mm} 80.5$^{\downarrow}$ &71.8 \\
\autorater{} &84.6 &\textbf{89.4} &86.2 &80.2 &83.8 &\textbf{62.7} &\textbf{82.3} &\textbf{86.5} &\textbf{72.6} \\
\bottomrule
\end{tabular}%
}
\caption{\textbf{ROC AUC scores on ImagenHub} for textual alignment, subject preservation, and their harmonic mean (as a unified evaluation) across \emph{Animal} and \emph{Object} categories, as well as for the \emph{Multi-subject} setting.}
\label{tab:imagenhub_results}
\end{table*}

\begin{table}[!t]
\centering
\resizebox{0.9\columnwidth}{!}{%
\begin{tabular}{lcccc}\toprule
&\makecell{\textbf{Textual} \\ \addlinespace[-2pt] \textbf{Alignment}} &\makecell{\textbf{Subject} \\ \addlinespace[-2pt] \textbf{Preservation}} &\makecell{\textbf{Unified} \\ \addlinespace[-2pt] \textbf{Evaluation}} \\\midrule
CLIP &\hspace{1mm} 83.2$^{\downarrow}$ &80.1 &\hspace{1mm} 81.5$^{\downarrow}$ \\
DINO &- &85.4 &- \\
Crop-IR &- &\hspace{1mm} \textbf{90.2$^{\uparrow}$} &- \\
CLIPScore &\hspace{1mm} 83.3$^{\downarrow}$ &- &- \\
BLIPScore &\hspace{1mm} 82.6$^{\downarrow}$ &- &- \\
SigLIP &\hspace{1mm} 75.3$^{\downarrow}$ &- &- \\
TIFA &\hspace{1mm} 90.6$^{\downarrow}$ &- &- \\
VQAScore &\hspace{1mm} 89.0$^{\downarrow}$ &- &- \\
VIEScore &\hspace{1mm} 82.5$^{\downarrow}$ &87.5 &84.9 \\
DreamBench++ &\hspace{1mm} 87.0$^{\downarrow}$ &\hspace{1mm} 89.9$^{\uparrow}$ &88.4 \\
\hdashline
PaliGemma\textsubscript{text/ref} &94.5 &87.5 &\textbf{90.8} \\
\autorater{} &\textbf{97.0} &82.2 &88.9 \\
\bottomrule
\end{tabular}%
}
\caption{\textbf{ROC AUC scores on KITTEN (landmarks)} for textual alignment, subject preservation, and their harmonic mean (as a unified evaluation).}
\label{tab:kitten_results}
\end{table}

\section{Results}
\label{sec:results}

Our main results are summarized in Tables~\ref{tab:dreambench_results}, \ref{tab:imagenhub_results}, and \ref{tab:kitten_results}, with qualitative examples in Fig.~\ref{fig:qualitative_examples_text_and_image}.

On DreamBench++ (Table~\ref{tab:dreambench_results}), \autorater{} outperforms or statistically matches all baselines across both criteria, with a notable 6.3-point lead over the GPT-4o-based DreamBench++ metric in \textit{subject preservation} for \emph{Objects}. 
This is especially notable given the benchmark's diverse visual styles, including cartoonish and pixelated images, which are outside \autorater{}'s training distribution of real-world video frames.
Similarly, on ImagenHub (Table~\ref{tab:imagenhub_results}), \autorater{} matches or exceeds all baselines in both single- and multi-subject settings, with 5.9-point gains over the strongest non-finetuned model on \textit{subject preservation} of the multi-subject setting (Crop-IR).
Lastly, on KITTEN (Table~\ref{tab:kitten_results}), \autorater{} leads in \emph{textual alignment} but underperforms in \emph{subject preservation}, though it remains statistically comparable to most baselines. This may result from \autorater{}'s identity-sensitive training, which penalizes minor deviations---especially challenging for landmarks with intricate visual details (Fig.~\ref{fig:qualitative_examples_image_landmark})---and from a domain shift, as landmarks were absent from \autorater{}'s training data (OOD).

Notably, across all benchmarks, fine-tuning only for \emph{textual alignment} (PaliGemma\textsubscript{text}) slightly reduces performance, especially for \emph{Animals} and \emph{Humans}, while training solely for \emph{subject preservation} (PaliGemma\textsubscript{ref}) yields even larger declines---up to a 14.8 points for \emph{Humans} (Table~\ref{tab:dreambench_results}). 
This suggests that joint training provides complementary benefits, with subject preservation gaining the most.

Fig.~\ref{fig:qualitative_examples_text_and_image} further showcases \autorater{}'s strengths, like its sensitivity to subtle \emph{textual alignment} errors, such as a missing waterfall (top-left). For \emph{subject preservation}, it remains robust to identity-agnostic changes, like a zoomed-out parrot or person (top-center and bottom-center) or different clothes (bottom-center), while staying sensitive to key identity traits, e.g., changed facial features (bottom-left) and colors (left and middle balloons).


Overall, \autorater{} consistently outperforms or statistically matches all baselines on both criteria, with the only exception of \textit{subject preservation} in the OOD landmarks category, where it still performs competitively. Importantly, it offers the best trade-off between \emph{textual alignment} and \emph{subject preservation}, surpassing all non-finetuned metrics in \emph{Unified Evaluation} across all benchmarks.

\section{Applicability to Rare Entities}
\label{sec:unknown_entities}

To test \autorater{} on unfamiliar subjects, we use the ImageRAG benchmark \cite{shalev2025imagerag}, which evaluates generated images based on prompts and reference images of uncommon subjects (e.g., scientific animal names, lesser-known dishes). 
Human annotators compared image pairs, selecting the better one based on \emph{Textual Alignment}, \emph{Visual Quality} (evaluating general depiction of the \textit{entity} rather than exact reference-adherence), and \emph{Overall Preference}. 
We report per-axis accuracy, defined as the frequency with which a metric ranks the human-preferred image higher, with approximate ties handled by rounding. \emph{Overall Preference} is computed as the harmonic mean of textual and visual scores, and significance testing is assessed via bootstrap resampling, akin to \S\ref{subsec:meta_eval_and-benchmarks}.

\begin{figure*}[!t]
  \centering
  \includegraphics[width=0.8\linewidth]{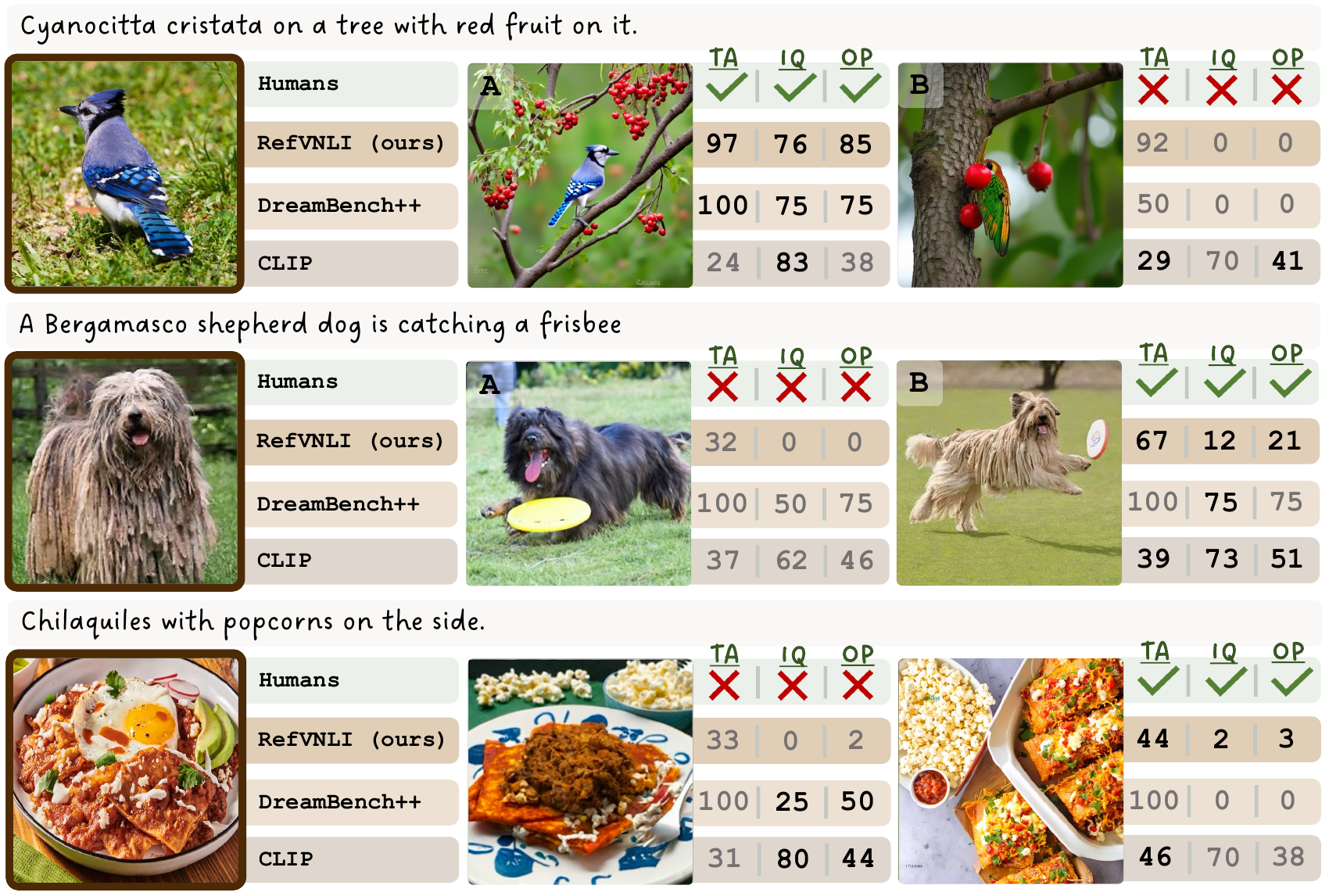}

   \caption{\textbf{ImageRAG Rare Entities Examples:} We compare \autorater{} with CLIP and DreamBench++ in aligning with human preferences (top rows of each example) across \textbf{Textual Alignment (TA)}, \textbf{Image Quality (IQ)}, and \textbf{Overall Preference (OP)}. 
   DreamBench++ scores (0–4) are rescaled to 0–100 for readability. The higher of the two criterion-wise scores is emphasized unless both are equal.}
   \label{fig:imagerag_cherrypicking}
\end{figure*}

As shown in Table~\ref{tab:rare_entities_imagerag_results}, \autorater{} consistently outperforms all baselines in aligning with human preferences across these criteria, showcasing strong robustness to rare subjects.\footnote{TIFA was excluded due to assigning identical scores to 61\% of pairs, making accuracy calculations unreliable.} This is further supported by Fig.~\ref{fig:imagerag_cherrypicking}, where only \autorater{} repeatedly matches human selections.

\begin{table}[!t]
\centering
\resizebox{0.9\columnwidth}{!}{%
\begin{tabular}{lcccc}\toprule
&\makecell{\textbf{Textual} \\ \addlinespace[-2pt] \textbf{Alignment}} &\makecell{\textbf{Visual} \\ \addlinespace[-2pt] \textbf{Quality}} &\makecell{\textbf{Overall} \\ \addlinespace[-2pt] \textbf{Preference}} \\\midrule
CLIP &\hspace{1mm} 51.8$^{\downarrow}$ &91.3 &\hspace{1mm} 69.2$^{\downarrow}$ \\
DINO &- &91.4 &- \\
Crop-IR &- &\hspace{1mm} 86.4$^{\downarrow}$ &- \\
CLIPScore &\hspace{1mm} 47.4$^{\downarrow}$ &- &- \\
BLIPScore &\hspace{1mm} 39.6$^{\downarrow}$ &- &- \\
SigLIP &\hspace{1mm} 74.8$^{\downarrow}$ &- &- \\
VQAScore &\hspace{1mm} 52.3$^{\downarrow}$ &- &- \\
VIEScore &\hspace{1mm} 60.8$^{\downarrow}$ &\hspace{1mm} 65.4$^{\downarrow}$ &\hspace{1mm} 69.6$^{\downarrow}$ \\
DreamBench++ &\hspace{1mm} 56.6$^{\downarrow}$ &\hspace{1mm} 83.1$^{\downarrow}$ &\hspace{1mm} 78.9$^{\downarrow}$ \\
\hdashline
PaliGemma\textsubscript{text/ref} &\hspace{1mm} 61.6$^{\downarrow}$ &\hspace{1mm} 83.0$^{\downarrow}$ &\hspace{1mm} 83.0$^{\downarrow}$ \\
\autorater{} &\textbf{87.2} &\textbf{95.5} &\textbf{91.4} \\
\bottomrule
\end{tabular}%
}
\caption{\textbf{Results on ImageRAG rare concepts}, where users select the better image in each pair based on textual alignment, visual quality (general entity depiction rather than specific subject adherence), and overall preference.
We report accuracy: how often models ranked the human-preferred image higher. Overall preference is the harmonic mean of textual and visual scores.}
\label{tab:rare_entities_imagerag_results}
\end{table}

\section{Related Work}
\label{sec:related_work}

Evaluation of Visual Language Models (VLMs) spans various settings, including visual reasoning \cite{bitton2024visual, kahou2017figureqa} and visual question-answering \cite{antol2015vqa, marino2019ok, mensink2023encyclopedic}. For text-to-image (T2I) models, assessments normally focus on image quality \cite{heusel2017gans, salimans2016improved}, diversity \cite{rassin2024grade}, and alignment with the text \cite{hessel2021clipscore, radford2021learning, hu2023tifa, yarom2023you, zhai2023sigmoid, lin2024evaluating}. 
Assessing subject preservation, which is crucial for subject-driven generation, is typically done using embedding-based metrics like CLIP \cite{radford2021learning} and DINO \cite{caron2021emerging}. Other metrics, like VIEScore \cite{ku-etal-2024-viescore} and DreamBench++ \cite{peng2024dreambenchhumanalignedbenchmarkpersonalized}, use GPT-4o \cite{openai2024gpt4technicalreport} to measure both criteria.


Subject-driven T2I models have been gaining much traction, with some methods fine-tuning general models into specialist versions that capture specific subjects and styles \cite{gal2022image, kumari2023multi, ruiz2023dreambooth, sohn2023styledrop, park2024cat}. 
Others focus on broader applicability using one-shot examples, either through adapter-based methods that integrate encoded reference images into diffusion models \cite{gal2023encoder, jia2023taming, wei2023elite, ye2023ip}  or via adapter-free techniques that directly use extracted features such as attention maps \cite{liu2023cones, hertz2024style, lv2024pick}.



Closely related, image editing complements subject-driven T2I generation in that the generated image's appearance is primarily governed by the input image, with the text only impacting specific aspects, whereas in out setting it is the other way around.
The task has evolved from pixel-to-pixel translation for predefined transformations \cite{isola2017image, zhu2017unpaired, wang2018high} to more flexible, text-guided edits \cite{brooks2023instructpix2pix, tumanyan2023plug, parmar2023zero}, with recent diffusion-based methods improving precision via cross-attention manipulation \cite{hertz2022prompt, yang2023dynamic}.
Beyond images, personalized generation extends to other modalities, including videos and texts. Video generation can be conditioned on text \cite{li2018video, hong2022cogvideo, singer2022make}, reference images \cite{wei2024dreamvideo, zhou2024sugar}, or other videos \cite{ku2024anyv2v}. In text generation, efforts focus on style transfer \cite{reif2022recipe, zhang2024distilling}, debiasing \cite{zhao2018gender, ravfogel2020null}, and broader semantic control \cite{shapira-etal-2022-interactive, slobodkin-etal-2023-summhelper, xie2023interactive}.

Finally, several studies leveraged intra-frame relationships in videos to learn more human-aligned visual representations. 
These works aim to improve robustness to identity-agnostic variations (e.g., rotation, lighting), by analyzing consecutive frames sourced from public video datasets \cite{jin2018unsupervised, parthasarathy2023self, wang2015unsupervised, wang2017transitive, wu2021contrastive} or captured by cameras on moving agents \cite{agrawal2015learning, jayaraman2015learning}.

\section{Conclusion}
\label{sec:conclusion}

We present \autorater{}, a cost-effective and reliable metric for subject-driven T2I evaluation that jointly assesses \emph{textual alignment} and \emph{subject preservation}. Trained on a large-scale, auto-generated dataset, \autorater{} is designed to be robust to identity-agnostic visual variations (e.g., pose, lighting, background) while remaining sensitive to identity-specific features (e.g., facial features, object shape, and unique details) when evaluating subject preservation. For textual alignment, it leverages subject-specific prompts with perturbed hard negatives to detect and penalize fine-grained mismatches. Across benchmarks, \autorater{} outperforms or rivals all baselines, including larger GPT-4o-based metrics, particularly on less-common subjects.

Future work should focus on improving performance across artistic styles as well as identity-altering edits, and supporting multiple reference images. More broadly, \autorater{} facilitates progress in personalized T2I generation by enabling better checkpoint selection, reinforcement learning, and iterative model refinement.\avivslobodkin{Added the last sentence to say that incorporating RefVNLI in downstream uses like better training and ckpt-selection is something we leave for future work (following the reviews). Is it sufficient?}


\section{Limitations}
\autorater{} was trained on data sourced from real-life video frames and images. While it performs well on stylistically consistent inputs, including cartoonish or pixelated images, it struggles with cross-style scenarios where \RefImageName{} and \TgtImageName{} differ in style, as well as when the subject undergoes explicit modifications (e.g., changes in color or shape). Additionally, the current framework is limited to single-reference cases and should be extended to support multiple references, both for the same subject and for distinct ones.

Moreover, research on subject-driven generation could benefit from a unified score capturing overall performance, rather than the two separate scores currently provided by \autorater{}. Although the harmonic mean of the two offers a reasonable proxy, future iterations should aim to output a single, integrated metric, alongside the individual, more granular scores.




\bibliography{custom}

\clearpage

\appendix

\section{Reproducibility}
\label{appendix:Reproducibility}
\begin{figure*}[t]
  \centering
  \includegraphics[width=1\linewidth]{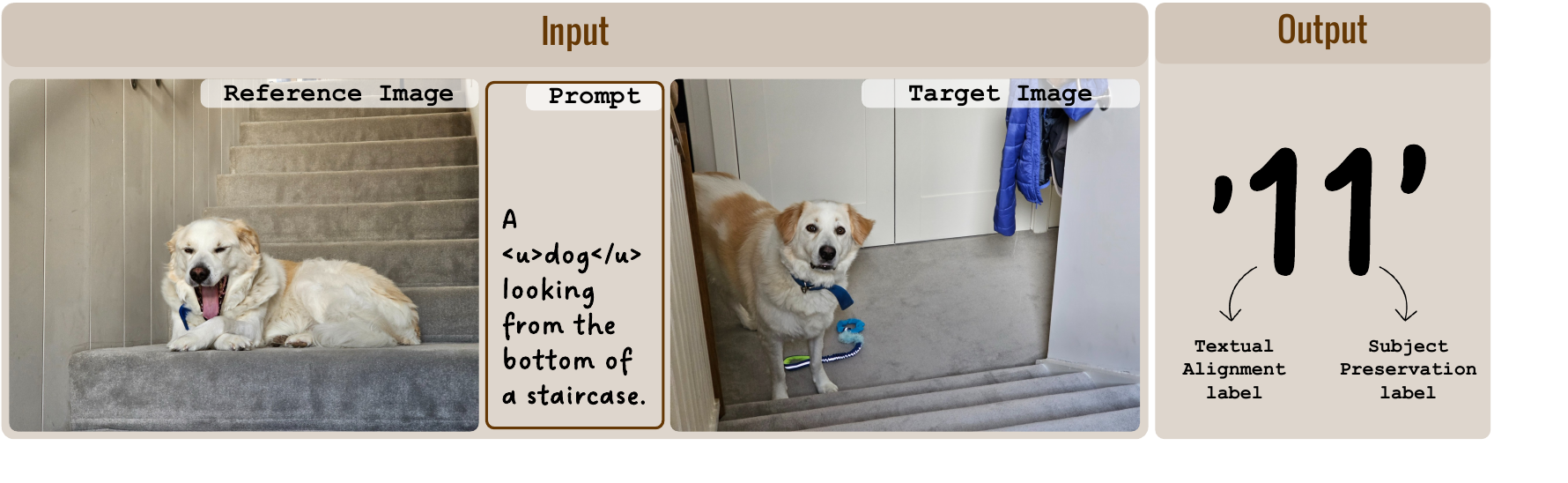}

   \caption{Example of the input and output of \autorater{}. The input consists of \RefImageName{} and \TgtImageName{}, as well as from the \TextDescName{} with \textit{$<$u$>$} and \textit{$<$\textbackslash{}u$>$} markups around the referenced subject (e.g., dog), while the output consists of two digits of \textit{`0'} or \textit{`1'}, with the first digit representing the first and second digits being the \textit{textual alignment} and \textit{subject preservation} labels, respectively.}
   \label{fig:output_input_example}
\end{figure*}

We fine-tuned PaliGemma \cite{beyer2024paligemma} on a balanced subset of our dataset, created by undersampling the more frequent labels. Training was conducted over 24 hours using two NVIDIA A100 GPUs (80 GiB each) with a batch size of 4. During training, we enclosed the referenced subject in the \TextDescName{} with \textit{$<$u$>$} and \textit{$<$\textbackslash{}u$>$} markups, and provided it alongside the separately passed \RefImageName{} and \TgtImageName{}. The model was trained to generate one of four strings---\textit{`00'}, \textit{`01'}, \textit{`10'}, \textit{`11'}---where the first and second digits represent \textit{textual alignment} and \textit{subject preservation}, respectively. See Fig.~\ref{fig:output_input_example} for an input-output example.

At inference, we used the same input structure, including the insertion of markups around the subject in the \TextDescName{} and separate image inputs. The \textit{textual alignment} and \textit{subject preservation} scores were then computed from the probabilities of the first and second generated tokens, respectively, with each score defined as the probability of token \textit{`1'} divided by the sum of probabilities for tokens \textit{`0'} and \textit{`1'}.

\section{Additional Details on the Meta-evaluation}
\label{appendix:meta_eval}
For meta-evaluation, we perform bootstrapping \citep{efron1987better} (1,000 samples per benchmark, with repetitions). For the ImageRAG rare entities benchmark, consisting of only 26 instances, we sample each time a sample 4 times this size, to ensure there are at least 100 instances. For the other benchmarks, which are significantly larger, the size of each sample is identical to the original size of the corresponding benchmark. For the calculations of confidence intervals (CIs), we use significance level $p=0.05$.

\begin{figure*}
\promptfont
\centering
\begin{tabular}{p{0.9\linewidth}}
\toprule
Context \\
\midrule
\textbf{Misalignment Injection Instructions (Short Captions)}
\begin{enumerate}
    \item \textbf{Understand the Caption:} Carefully read the short caption to fully grasp the scene it describes.
    \item \textbf{Identify and Swap:} Select a single visual detail within the caption to modify. Replace this detail with a \textit{different, incorrect, but still plausible} visual detail. For example, you might change a color, an object, or a location. \textbf{Do not modify the underlined entity (if any).}
    \item \textbf{Apply the Tags:} Enclose the \textit{original} visual detail within \texttt{<swap>} tags. Immediately \textit{after} the closing \texttt{</swap>} tag, write the \textit{new, incorrect} visual detail.  There should be no space between the closing \texttt{</swap>} and the new word.

       Example: If the original sentence is "The cat sat on the red mat," and you want to change "red" to "blue," the result should be: "The cat sat on the \texttt{<swap>red</swap><blue>} mat."
    \item \textbf{Final Check:} Ensure the modified caption is grammatically correct and reads naturally, even though it now contains a factual error. The sentence should be internally logical, despite contradicting the actual visual content. Again, \textbf{ensure the underlined entity (if any) remains completely unchanged.} 
\end{enumerate}\\
\midrule
Few-Shot \\
\midrule
\textbf{Here are some examples:}\\
\\
\textbf{INPUT:} A woman is sitting in a living room, and \texttt{<u>}she\texttt{</u>} is looking at something with a concerned expression\\
\textbf{OUTPUT:} A woman is sitting in a \texttt{</swap>}living room\texttt{</swap><}kitchen\texttt{>}, and \texttt{<u>}she\texttt{</u>} is looking at something with a concerned expression.\\
\\
\textbf{INPUT:} Two men are sitting on a leather couch in a living room. One \texttt{<u>}man\texttt{</u>} is sitting on the left side of the couch, looking at a laptop. The other man is sitting on the right side of the couch, talking on a phone. The room is decorated with various items, including a large model of a spaceship.\\
\textbf{OUTPUT:} Two men are sitting on a leather couch in a living room. One \texttt{<u>}man\texttt{</u>} is sitting on the left side of the couch, looking at a laptop. The other man is sitting on the right side of the couch, talking on a phone. The room is decorated with various items, including a large model of a \texttt{<swap>}spaceship\texttt{</swap><}sailboat\texttt{>}.\\
\\

Now it's your turn! Follow the instructions. Answer only with the corrupted sentence, Don't forget to add the tags.\\
\\
\textbf{INPUT:} A lizard is perched on a rock, surrounded by other rocks and foliage. The \texttt{<u>lizard</u>} is facing the camera, with its head raised and its tail curled behind it. \\
\\
\textbf{OUTPUT:} \\
\midrule
Generated \\
\midrule
\textbf{OUTPUT:} A lizard is perched on a \textbf{\texttt{<swap>rock</swap><branch>}}, surrounded by other rocks and foliage. The \texttt{<u>lizard</u>} is facing the camera, with its head raised and its tail curled behind it.\\

\bottomrule
\end{tabular}   
\caption{Hard Negative Caption Generation.  This figure illustrates the prompting strategy used to generate hard negative captions, containing a single, plausible but factually incorrect visual detail, for enhanced misalignment detection.}
\label{fig:corruption_prompt}

\end{figure*}

\section{Further Information on the Data Construction Pipeline}
\label{appendix:more_on_data_construction}

\subsection{Collection of Subject-driven Image Pairs}
\label{subappendix:collection_of_subject_drive_image_pairs}
To reduce noise when collecting subject-driven image pairs, we applied several filtering steps, including the removal of blurred images and those not depicting the intended subject. Subject presence was verified using Gemini \citep{geminiteam2024geminifamilyhighlycapable} (version \textit{gemini-2.0-flash}). For the subset sourced from TVQA+, we additionally filtered out frames containing subtitles or credits, also using Gemini.

For identity-sensitive image pairs, we used Stable Diffusion \citep{Rombach_2022_CVPR} for inpainting,\footnote{\url{https://huggingface.co/stabilityai/stable-diffusion-2-inpainting}} with $\eta = 1.0$ and a guidance scale of 3.0. We retained only images with a full mask size of at least 60,000 pixels (20,000 for humans, focusing on facial regions). Five patches of 250–300 pixels were randomly sampled and inpainted. To increase the likelihood of meaningful subject changes, we further filtered out inpainted images where all patchwise MSE values fell below 6,500 for objects, 5,400 for animals, and 20,000 for humans.

\subsection{Collection of Image-\TextDescName{} Pairs}
\label{subappendix:collection_of_subject_drive_image_pairs}
For the image-captioning of the \TgtImageName{} with the inserted bounding boxes, we employed Gemini \citep{geminiteam2024geminifamilyhighlycapable} (version \textit{gemini-2.0-flash}).



\subsection{Generation of \TextDescName{}-\TgtImageName{} Hard-negatives}
Fig.~\ref{fig:corruption_prompt} showcases the prompt used to generated the \TextDescName{}-\TgtImageName{} hard negatives.

\begin{table*}[!htp]
\centering
\resizebox{0.7\textwidth}{!}{%
\begin{tabular}{lccc@{\hskip 20pt}ccc@{\hskip 20pt}cccc}\toprule
&\multicolumn{3}{c}{\textbf{Textual Alignment}}
{\hskip -15pt}
&\multicolumn{3}{c}{\textbf{Subject Preservation}}
{\hskip -15pt}
&\multicolumn{3}{c}{\textbf{Unified Evaluation}}
{\hskip -20pt}
\\ \addlinespace[1pt]
&DreamBench++ &\makecell{ImagenHub \\ \addlinespace[-1pt] {\small (Single/Multi)}} &KITTEN &DreamBench++ &\makecell{ImagenHub \\ \addlinespace[-1pt] {\small (Single/Multi)}} &KITTEN &DreamBench++ &\makecell{ImagenHub \\ \addlinespace[-1pt] {\small (Single/Multi)}} &KITTEN \\\midrule
\autorater{} (ours) &81.5 &87.7 / 86.3 &97.0 &82.7 &83.0 / 62.8 &82.3 &82.1 &85.3 / 72.7 &89.0 \\
\hspace{3mm} {\small reverse classification order} &80.0 &85.2 / 85.5 &95.3 &80.9 &84.3 / 68.7 &87.0 &80.4 &84.7 / 76.2 &91.0 \\
\hspace{3mm} {\small multiclass} &79.5 &83.7 / 84.7 &94.7 &79.6 &76.0 / 61.1 &86.3 &79.5 &79.7 / 71.0 &90.3 \\
\hspace{3mm} {\small separate classification} &79.7 &85.2 / 87.5 &95.8 &78.3 &77.1 / 56.7 &89.2 &79.0 &80.9 / 68.8 &92.4 \\
\hspace{3mm} {\small no markup} &78.4 &87.0 / 84.3 &92.3 &65.5 &75.9 / 60.8 &88.7 &71.4 &81.1 / 70.6 &90.5 \\
\hspace{3mm} {\small concatenated images} &79.6 &86.2 / 86.2 &93.6 &74.2 &81.1 / 81.2 &89.9 &76.8 &83.6 / 83.6 &91.7 \\

\hspace{3mm} {\small only video-based training} &80.4 &85.2 / 83.8 &96.3 &77.6 &77.5 / 66.4 &89.7 &79.0 &81.2 / 74.1 &92.9 \\
\bottomrule
\end{tabular}%
}
\caption{\textbf{Ablation Study:}
ROC AUC scores for various ablated versions of \autorater{} across benchmarks (over all subjects). 
The ablations evaluate alternative output formulations, such as reversed classification order, a four-label multiclass framework, and separate aspect classification via a special token. Additional variants exclude subject markup from input prompt, merge reference and target images, or remove identity-sensitive training examples by using only video-based instances.}
\label{tab:ablations_full}
\end{table*}

\section{Ablations}
\label{appendix:ablations}

To assess the impact of various design decisions in \autorater{}, we run an ablation study examining alternative input and output configurations. On the output side, we test: reversing the classification order (subject preservation before textual alignment); a 4-label multiclass framework for joint text-image alignment classification; and a model that prefixes a designated token (\textit{`TEXT'} or \textit{`IMAGE'}) to the \TextDescName{} to enable separate classification of each aspect within a unified model. For inputs, we explore the effect of removing subject markup from the \TextDescName{} and of concatenating \RefImageName{} and \TgtImageName{} instead of passing them separately. Finally, we also explore the impact of omitting the identity-sensitive training examples (see \S\ref{subsec:trainset_construction_subject_consistency}), by only using the video-based instances during training.

Results (Table~\ref{tab:ablations_full}) show that reversing the classification order degrades performance, particularly in subject preservation, as does evaluating each aspect separately. 
This suggests that first evaluating textual alignment helps in subject preservation assessment. 
The multiclass approach also underperforms compared to our dual binary classification setup, highlighting the benefits of treating each criterion independently. Further, removing subject markup weakens subject preservation, underscoring its role in linking the reference image to the \TextDescName{}. Additionally, concatenating images instead of processing them separately harms performance, emphasizing the advantage of distinct image inputs. Finally, excluding identity-sensitive training instances leads to notable drops in subject preservation, underscoring their importance.


\begin{table}[!htp]
\centering
\resizebox{\columnwidth}{!}{%
\begin{tabular}{lcccccccc}\toprule
&\multicolumn{3}{c}{\textbf{DreamBench++}} &\multicolumn{3}{c}{\textbf{ImagenHub}} &\textbf{KITTEN} \\
&Animal &Human &Object &Animal &Object &Multi-subj. &Landmarks \\\midrule
OWL-ViT &77.4 &\textbf{87.1} &81.8 &74.8 &83.1 &52.0 &85.6 \\
GroundingDINO &\textbf{76.9} &85.6 &\textbf{83.4} &\textbf{77.6} &\textbf{84.1} &\textbf{56.8} &\textbf{90.2} \\
\bottomrule
\end{tabular}%
}
\caption{\textbf{ROC AUC scores of Crop-IR for subject preservation} when employed with two different object-detection models: OWL-ViT and GroundingDino. Bold indicates the highest score per column.}
\label{tab:crop_ir_object_detection_ablation}
\end{table}



\section{Crop-IR Object Detection Model Ablation}
\label{appendix:crop_ir_object_detection_ablation}
Deploying the Crop-IR metric \citep{winter2024objectmate} requires an object detection model to locate and crop the referenced subjects. To this end, we compare two prominent object detection models: OWL-ViT \citep{minderer2022simple}\footnote{\url{https://huggingface.co/google/owlv2-base-patch16}} and GroundingDino \citep{liu2024groundingdinomarryingdino}.\footnote{\url{https://huggingface.co/IDEA-Research/grounding-dino-base}} As shown in Table~\ref{tab:crop_ir_object_detection_ablation}, GroundingDino leads to better evaluation of subject preservation, and is therefore adopted to ensure a fairer evaluation.

\begin{table}[!t]
\centering
\resizebox{\columnwidth}{!}{%
\begin{tabular}{lcccc}\toprule
&\makecell{\textbf{Inference Time} \\ \addlinespace[-2pt] \textbf{(seconds)}} &\makecell{\textbf{GPU Memory Usage} \\ \addlinespace[-2pt] \textbf{(GiB)}} &\makecell{\textbf{API Calls Cost} \\ \addlinespace[-2pt] \textbf{(\$)}} \\\midrule
CLIP &0.1 &1.2 &- \\
DINO &0.06 &0.7 &- \\
Crop-IR &0.6 &5.8 &- \\
ArcFace &1.2 &- &- \\
CLIPScore &0.07 &0.6 &- \\
BLIPScore &0.7 &4.5 &- \\
SigLIP &0.03 &1.4 &- \\
TIFA &22.5 &26.4 &- \\
VQAScore &0.2 &23.1 &- \\
VIEScore &6.9 &- &0.04 \\
DreamBench++ (text) &3.2 &- &0.02 \\
DreamBench++ (ref) &1.0 &- &0.03 \\
\hdashline
PaliGemma\textsubscript{text} &0.4 &12.5 &- \\
PaliGemma\textsubscript{ref} &0.4 &12.5 &- \\
\autorater{} &0.5 &12.5 &- \\
\bottomrule
\end{tabular}%
}
\caption{Computational costs for all baseline models and \autorater{}, including per-instance inference time (in seconds), GPU memory usage (in GiB), and GPT-4o API costs (in \$, only when applicable), averaged across benchmarks. For DreamBench++, we report separate values for each evaluation criterion, as each requires a distinct API call under its framework. The final three rows present models fine-tuned on our dataset, with PaliGemma\textsubscript{text} and PaliGemma\textsubscript{ref} being the variants tuned exclusively for evaluating textual alignment and subject preservation, respectively.}
\label{tab:computational_cost}
\end{table}


\section{Computational Cost}
\label{appendix:computational_cost}
Table~\ref{tab:computational_cost} presents the computational costs of all baseline models and \autorater{}, in terms of inference time, GPU memory usage and GPT-4o API costs. 

\section{Additional Qualitative Examples for Subject Preservation Evaluation}
\label{appendix:subject_preservation_additional_examples}
\begin{figure*}[t]
  \centering
  \includegraphics[width=1\linewidth]{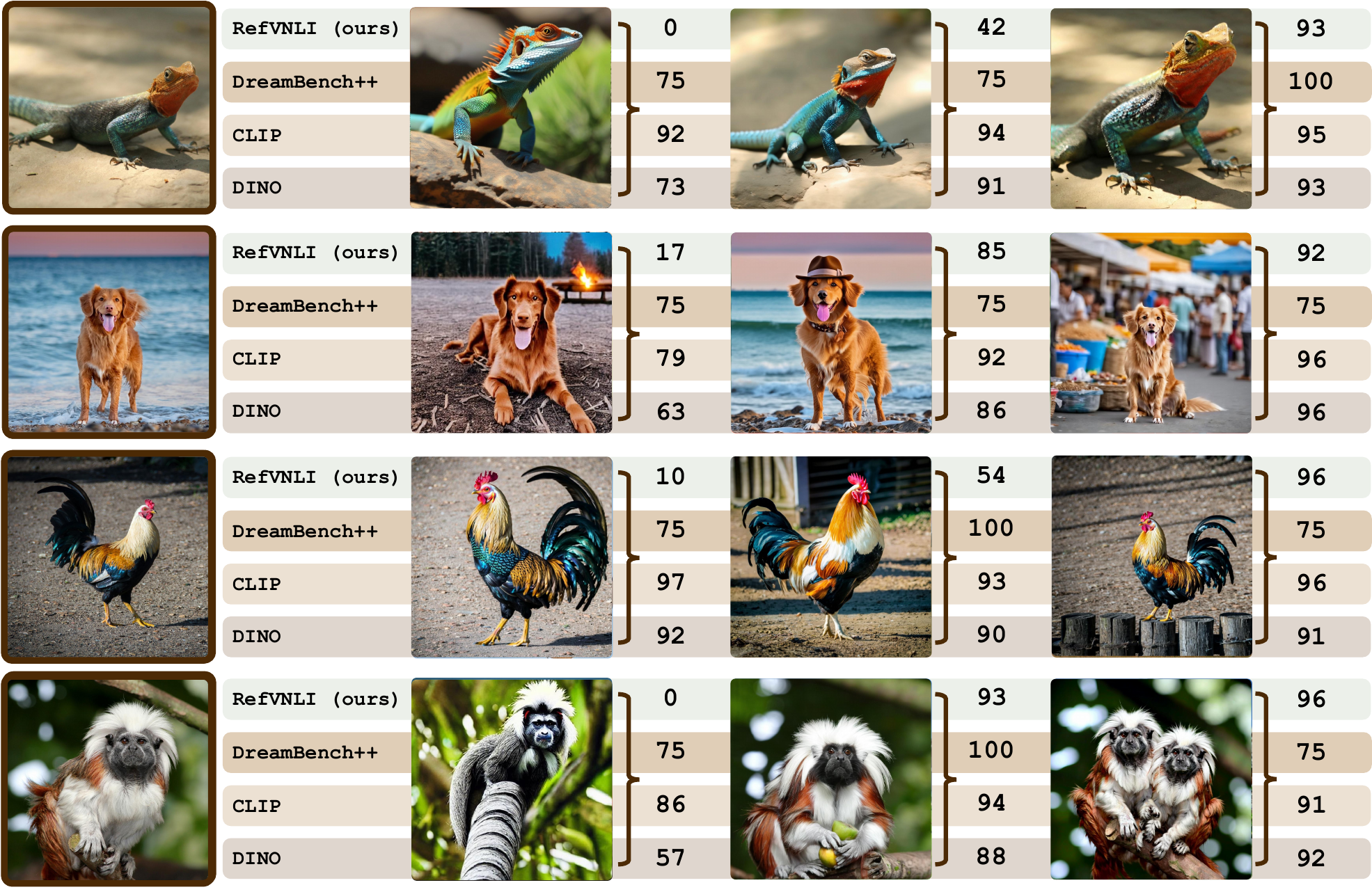}

   \caption{\textbf{Qualitative Examples of Subject Preservation Evaluation for the \emph{Animal} Category.} DreamBench++ scores (0-4) are scaled to 0-100 for better readability.}
   \label{fig:qualitative_examples_image_animal}
\end{figure*}

\begin{figure*}[t]
  \centering
  \includegraphics[width=1\linewidth]{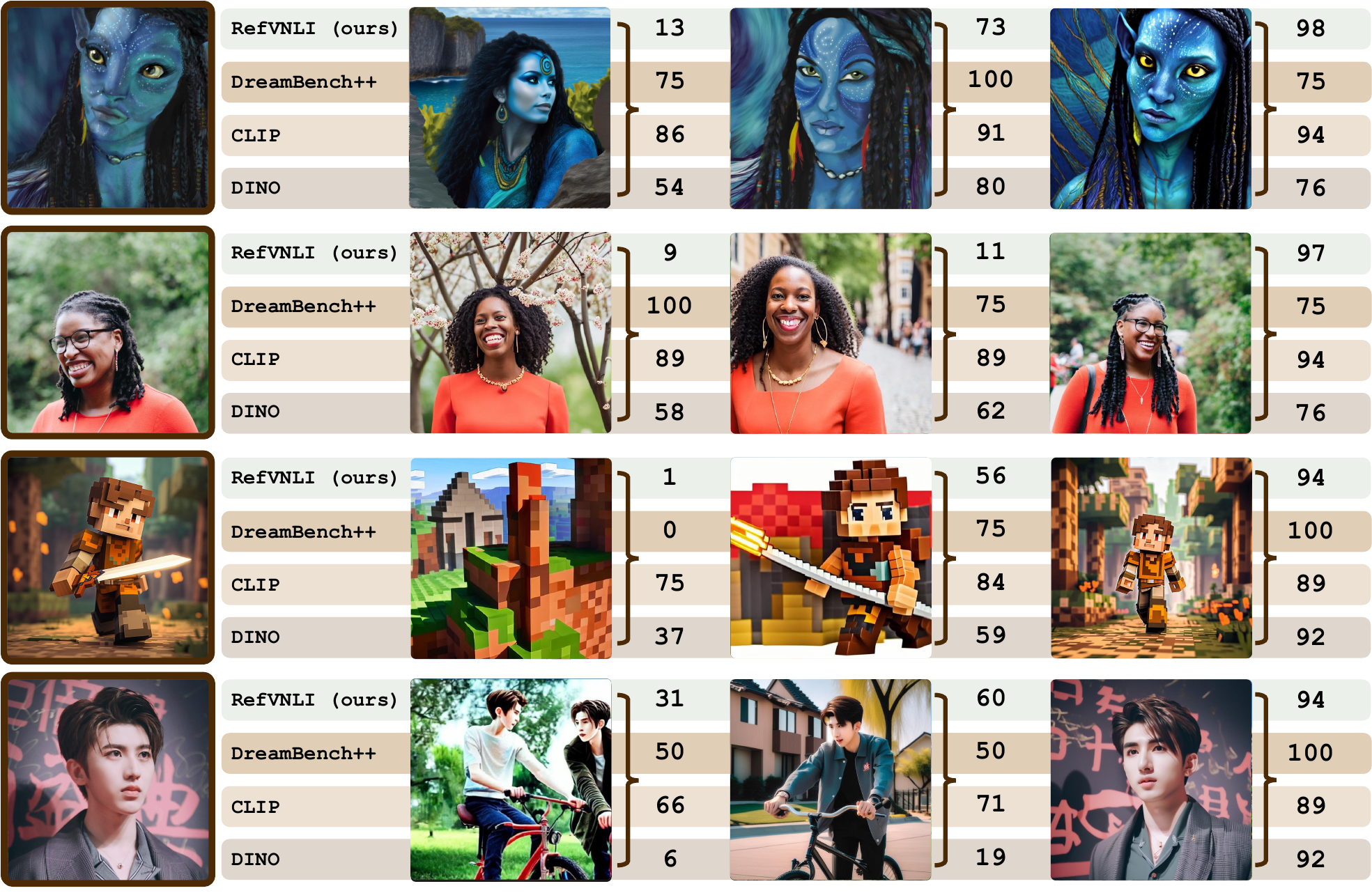}

   \caption{\textbf{Qualitative Examples of Subject Preservation Evaluation for the \emph{Human} Category.} DreamBench++ scores (0-4) are scaled to 0-100 for better readability.}
   \label{fig:qualitative_examples_image_human}
\end{figure*}

\begin{figure*}[t]
  \centering
  \includegraphics[width=1\linewidth]{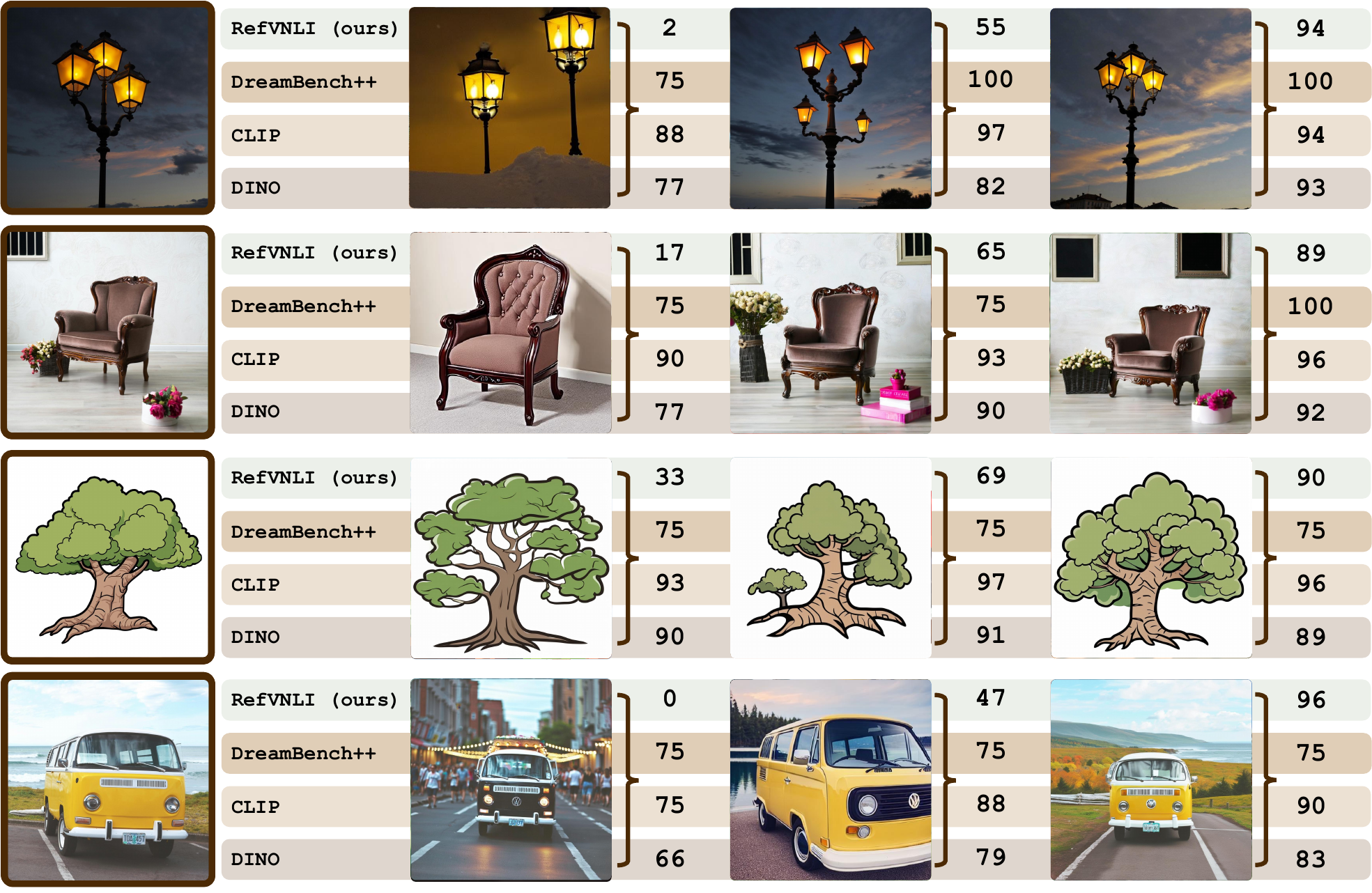}

   \caption{\textbf{Qualitative Examples of Subject Preservation Evaluation for the \emph{Object} Category.} DreamBench++ scores (0-4) are scaled to 0-100 for better readability.}
   \label{fig:qualitative_examples_image_object}
\end{figure*}

\begin{figure*}[t]
  \centering
  \includegraphics[width=1\linewidth]{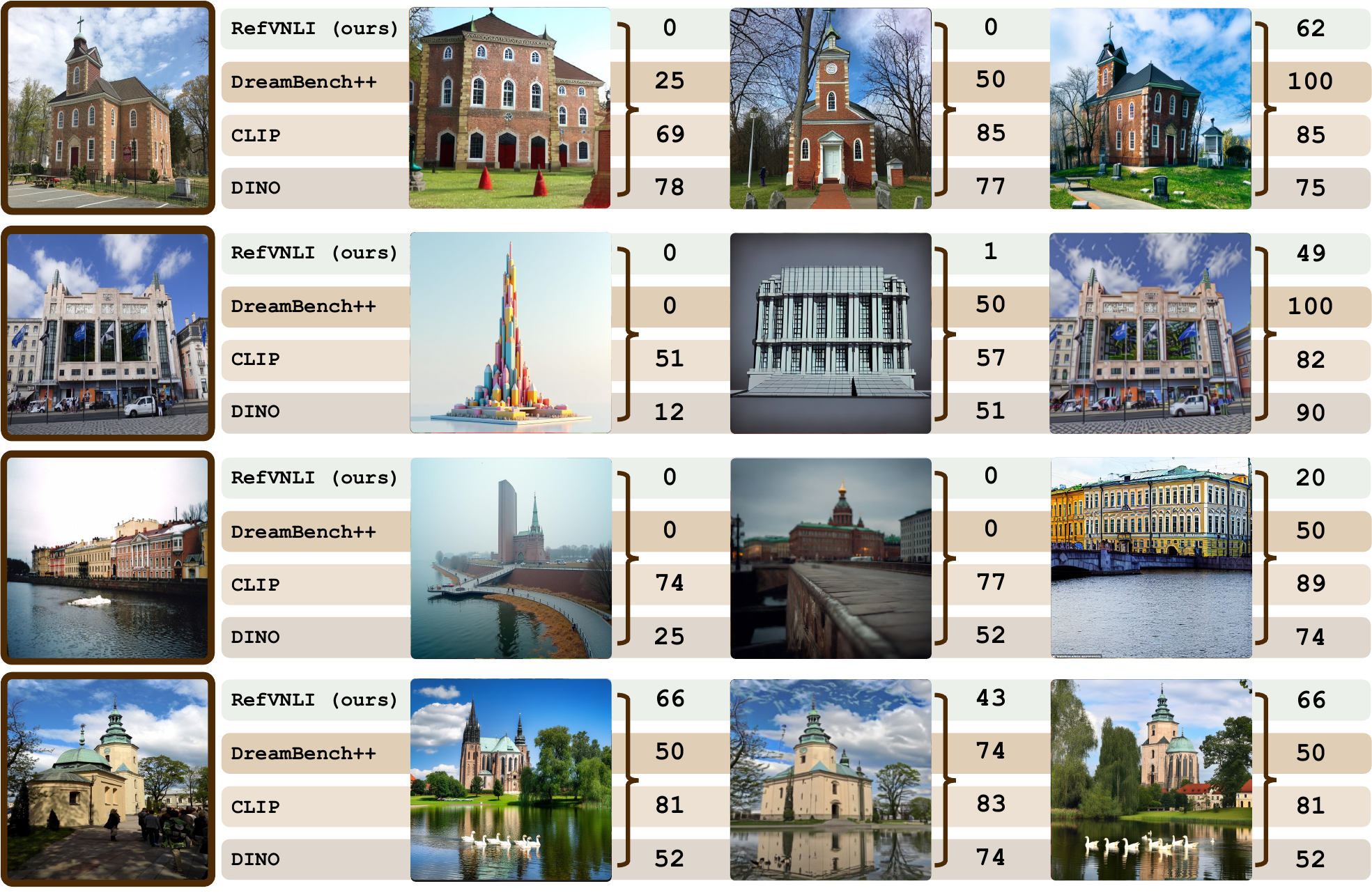}

   \caption{\textbf{Qualitative Examples of Subject Preservation Evaluation for the \emph{Landmark} Category.} DreamBench++ scores (0-4) are scaled to 0-100 for better readability.}
   \label{fig:qualitative_examples_image_landmark}
\end{figure*}

In Figures~\ref{fig:qualitative_examples_image_animal}, \ref{fig:qualitative_examples_image_human}, \ref{fig:qualitative_examples_image_object}, and \ref{fig:qualitative_examples_image_landmark} we present additional qualitative examples of \emph{subject preservation} evaluation for the \emph{Animal}, \emph{Human}, \emph{Object}, and \emph{Landmark} categories, respectively.

\section{List of Data and Software Licenses Employed in this Paper}
Our framework  dependencies are:
\begin{enumerate}
    \item Mementos dataset: \url{https://github.com/si0wang/Mementos}, Misc.
    \item TVQA+ dataset: \url{https://github.com/jayleicn/TVQAplus/blob/master/LICENSE}, under the MIT License.
    \item Open Images dataset: \url{https://github.com/openimages/dataset/blob/main/LICENSE}, under an Apache License 2.0.
    \item PaliGemma model: \url{https://ai.google.dev/gemma/terms}, under Gemma Terms of Use License.
    \item Gemini model: \url{https://ai.google.dev/gemini-api/docs/models}, under an Apache License 2.0.
    \item GPT-4o model: \url{https://github.com/openai/openai-openapi/blob/master/LICENSE}, under the MIT License.
    \item OWL-ViT model: \url{https://huggingface.co/google/owlv2-base-patch16}, under an Apache License 2.0.
    \item GroundingDino model: \url{https://huggingface.co/IDEA-Research/grounding-dino-base}, under an Apache License 2.0.
    \item Stable Diffusion inpainting model: \url{https://huggingface.co/stabilityai/stable-diffusion-xl-base-1.0/blob/main/LICENSE.md}, under Stability AI CreativeML Open RAIL++-M License.
\end{enumerate}

\end{document}